# Advancing Anomaly Detection: Non-Semantic Financial Data Encoding with LLMs

Alexander Bakumenko*
Clemson University
Charleston, SC, USA
alexander.bakumenko@outlook.com

Kateřina Hlaváčková-Schindler
University of Vienna
Vienna, Austria
katerina.schindlerova@univie.ac.at

Claudia Plant
University of Vienna
Vienna, Austria
claudia.plant@univie.ac.at

Nina C. Hubig
Clemson University
Charleston, SC, USA
nhubig@clemson.edu

## ABSTRACT

Detecting anomalies in general ledger data is of utmost importance to ensure trustworthiness of financial records. Financial audits increasingly rely on machine learning (ML) algorithms to identify irregular or potentially fraudulent journal entries, each characterized by a varying number of transactions. In machine learning, heterogeneity in feature dimensions adds significant complexity to data analysis. In this paper, we introduce a novel approach to anomaly detection in financial data using Large Language Models (LLMs) embeddings. To encode non-semantic categorical data from real-world financial records, we tested 3 pre-trained general purpose sentence-transformer models. For the downstream classification task, we implemented and evaluated 5 optimized ML models including Logistic Regression, Random Forest, Gradient Boosting Machines, Support Vector Machines, and Neural Networks. Our experiments demonstrate that LLMs contribute valuable information to anomaly detection as our models outperform the baselines, in selected settings even by a large margin. The findings further underscore the effectiveness of LLMs in enhancing anomaly detection in financial journal entries, particularly by tackling feature sparsity. We discuss a promising perspective on using LLMs' embeddings for non-semantic data in the financial context and beyond.

## 1 INTRODUCTION

### 1.1 Anomaly Detection in Finance and the Critical Role of Machine Learning

Financial data, sourced from the general ledger, details an organization's financial transactions including revenue, expenses, assets, and liabilities, serving to accurately document business activities [21, 24]. It is essential for ensuring compliance and transparency for stakeholders like regulatory bodies, investors, and financial institutions. Furthermore, the data supports decision-making through analytics, while anomaly detection is crucial for maintaining data integrity and reliability [37].

*Author to whom correspondence should be addressed.

Financial data anomalies, caused by errors, fraud, or manipulation, risk considerable financial loss, undermine investor confidence, and necessitate fraud investigations [24]. Prompt, effective anomaly detection is essential for regulatory adherence and financial protection [8, 22]. Traditional financial anomaly detection has struggled with complex, voluminous data and advancing fraud techniques. Early approaches, relying on manual checks and rule-based systems, were inefficient, missing many anomalies and producing numerous false alarms, allowing financial fraud to go unnoticed [5].

Machine learning (ML) has become crucial in modern financial auditing, enabling efficient processing and pattern recognition in large datasets. However, as financial fraud techniques grow more complex, there is a need for novel methods to overcome challenges in data preprocessing and model limitations [2, 22]. Recent research highlights the potential of Autoencoders for anomaly detection and LLMs for diverse tasks, yet the sparsity and complexity of real-world data limits their effectiveness [26, 36].

### 1.2 Problem Statement

This work addresses the challenge of efficient detecting anomalies in general ledger data, focusing on the issues of feature dimension heterogeneity and feature sparsity, which hinder anomaly detection in financial audits. The approach involves using sentence-BERT LLMs' pre-trained representations to encode non-semantic categorical data in journal entries, enhancing the identification of data irregularities. While current solutions use various vectorization methods with the subsequent dimensionality reduction of sparse features, this can be inadequate for real-world journal entries, which often differ greatly in length and complexity. Moreover, a non-temporal aspect of transaction anomalies limits the range of applicable feature encoding methods. Consequently, ML algorithms for detecting anomalies in financial data face difficulties with heterogeneous and sparse data, causing problems in encoding and classification, and thus, leading to suboptimal outcomes. This affects the reliability of financial records audit.

We propose a novel approach, conceptualized in Figure 1, using pre-trained sentence-transformer models to encode non-semantic financial data, addressing feature heterogeneity and sparsity issues. This approach diverges from conventional ML techniques in financial anomaly detection, suggesting a hybrid model that combines sentence-transformer embeddings with ML classifiers for enhanced anomaly detection performance. Accurate anomaly detection is a cornerstone for reliable financial audits. Improvements in handling heterogeneity and sparsity in financial data can significantly enhance anomaly detection processes, contributing to better risk management and regulatory compliance. The proposed approach has broader implications beyond financial anomaly detection, offering a template for applying advanced encoding techniques for complex datasets across domains.

### 1.3 Objectives and Contributions

In this work we formulate the following 2 hypotheses:

- **Hypothesis 1:** *Utilizing sentence-transformer LLMs for encoding non-semantic categorical data from financial records effectively standardizes feature variability enhancing the compactness and information retention of feature sets, when compared to conventional methods, measurable through dimensionality reduction techniques like PCA.*
- **Hypothesis 2:** *The integration of sentence-transformer based LLM embeddings with optimized ML models yields superior anomaly detection performance in financial journal entries, evidenced by improved evaluation metrics, compared to traditional ML approaches.*

In formulating our hypotheses, we draw on recent findings that demonstrate LLMs' adaptability beyond text-based tasks [28]. Studies have shown that LLMs, originally trained on text, can effectively process and encode non-textual, linguistically non-semantic data [30]. This capability, arising from the encoding functions of their transformer blocks, prompts our hypothesis 1, suggesting the use of the SBERT LLMs in transforming non-semantic financial datasets into standardized single-size vector features. Subsequently, our hypothesis 2 builds on the LLMs' efficient encoding capabilities, implying the integration of LLMs' embeddings could enhance ML models, particularly in detecting anomalies in financial data. Validating Hypothesis 1 would demonstrate a novel method to manage feature variability in financial records, enchancing anomaly detection. Confirming Hypothesis 2 would illustrate the effectiveness of integrating LLM embeddings with optimized models in detecting financial anomalies, potentially surpassing traditional methods.

Together, these findings could transform current practices in financial anomaly detection. The innovative use of LLMs could greatly advance this field, showing their potential for cross-disciplinary applications and improving financial auditing and monitoring systems.

## 2 BACKGROUND AND RELATED WORK

### 2.1 Machine Learning Methods and Limitations in Detecting Financial Anomalies

Anomaly detection in finance is critical, with fraudulent activities greatly affecting the sector. The rise of digital financial services, especially post-COVID-19 pandemic, necessitates advanced fraud detection methods [43]. Deep learning, including variational autoencoders and LSTM architectures, has shown success in detecting anomalies in journal entries [44] and e-commerce [27], with LSTMs also being effective [1]. Graph Neural Networks (GNNs) are notable for their ability to handle complex data relationships in fraud detection [43]. Various ML techniques, such as Naive Bayes, Logistic Regression, KNN, Random Forest, and Sequential CNNs, have been applied to credit card fraud detection [29], with CatBoost-based methods highlighting the role of feature engineering and memory compression in improving efficiency [13]. ML in finance is largely applied, from detecting journal entry anomalies to identifying fraudulent transactions in healthcare and banking [29, 38]. While case studies affirm their effectiveness, they also point out challenges in practical implementation [7].

Applying ML in financial fraud detection faces challenges due to evolving fraud techniques and the complexity of financial data [11]. Accurate modeling relies on high-quality, standardized data, as it was discussed in the context of the credit card industry [31]. Financial data's non-stationarity, non-linearity, and low signal-to-noise ratio complicate model training and performance [40], necessitating advanced methods for preprocessing complex data improving data quality and model performance. Enhancing data representation and simplifying features can also improve ML model interpretability, meeting the regulatory and compliance demands in finance. [38]. Additionally, balancing computational complexity with high detection accuracy is crucial [27], underscoring the need to enhance the compactness and information retention of feature sets. Promising research directions warrant exploration of diverse ML approaches and hybrid applications, emphasizing the importance of innovative data pre-processing and adaptable ML methods to tackle data quality and model adaptability challenges [4].

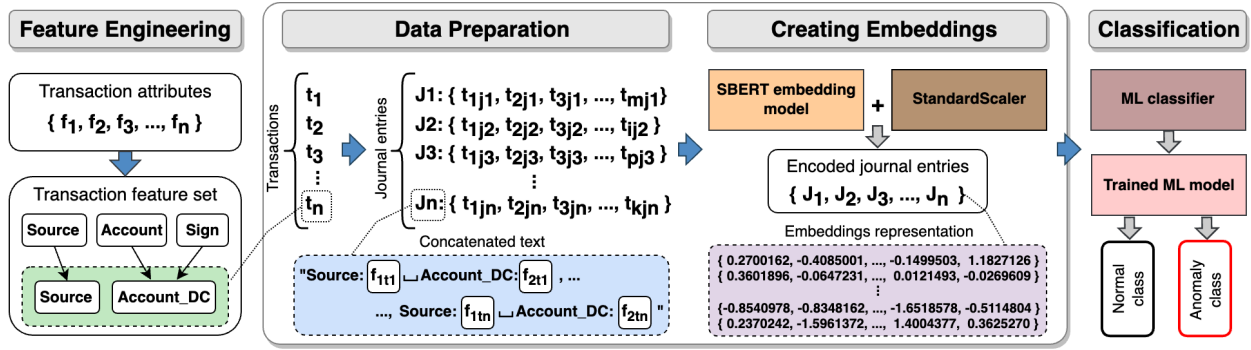

**Figure 1: A novel approach to efficiently encode journal entry non-semantic categorical features utilizing SBERT LLM model embeddings. It integrates *Data Preparation* and *Creating Embeddings* steps that produce one-size vectors feature set for the downstream ML tasks.**

## 2.2 Applications and capabilities of LLMs

LLMs such as GPT-3, PaLM, and LLaMA mark a paradigm shift in natural language processing (NLP) and artificial intelligence (AI), evolving from rule-based frameworks to sophisticated neural network architectures like Transformer. This evolution allows LLMs to encode vast linguistic datasets into vector representations for diverse applications [41, 42]. LLMs such as BERT excel at capturing the complex semantic and syntactic nuances of language, resulting in dense embeddings. These embeddings are pivotal for tasks like node classification in textual graphs [14], demonstrating LLMs' ability to generate meaningful representations from vast textual corpora [32]. Originally designed for linguistic tasks, LLMs show remarkable versatility by branching into non-linguistic domains, effectively encoding diverse data types, including non-semantic elements, into sequential formats. Illustratively, sentence-transformers vectorize non-linguistic data, extending LLMs' use to computer vision [30]. LLMs excel in tasks such as text summarization and content recommendation, thereby proving their broad applicability [25, 41, 42]. LLMs provide innovative approaches in data analysis by effectively managing feature variability and sparsity, thus enhancing anomaly detection. They can surpass traditional ML in processing complex data for advanced analyses [3, 20].

In financial analysis, LLMs constitute a major methodological leap. Sentence-transformers underscore the ability of LLMs to tackle feature heterogeneity and sparsity in anomaly detection by producing meaningful vectors [34, 39, 40]. Sentence-BERT (SBERT), a refined version of BERT, produces semantically dense sentence embeddings, improving clustering and semantic search [34]. SBERT leverages siamese and triplet networks to enhance sentence semantic analysis, ensuring similar sentences are close in embedding space. This improvement reduces embedding generation time from 65 hours with BERT to seconds for large datasets. SBERT excels in various tasks like sentence pair regression and semantic similarity, demonstrating its potential in fast, high-quality embeddings for both linguistic and non-linguistic data applications, beyond traditional text-based tasks [34].

## 2.3 Identifying Research Gaps in Financial Anomaly Detection

Despite progress in ML and deep learning for financial anomaly detection, these approaches often falter due to the diverse and sparse nature of financial data, particularly in journal entries, undermining the effectiveness of data encoding and classification, and consequently, the precision and reliability of financial audits [6]. Traditional anomaly detection techniques rely on vectorization and dimensionality reduction, but these may not suffice for real-world journal entries, which greatly differ in length and complexity. Moreover, the non-temporal aspect of financial transactions restricts the use of some feature encoding strategies.

Advanced ML techniques remain underleveraged for non-semantic, categorical financial data, with traditional anomaly detection methods falling short in addressing the non-temporal and heterogeneous data complexities. The untapped potential of sentence-transformer LLMs for financial data analysis presents an opportunity to innovate in handling feature variability and sparsity. Bridging the significant research gap by integrating LLM embeddings with optimized ML models for financial anomaly detection could significantly enhance classification accuracy and data encoding robustness, outperforming traditional methodologies.

## 3 DATA DESCRIPTION AND ETHICAL CONSIDERATIONS

In our work, we utilized an aggregated real-world General Ledger dataset from various anonymized companies, as described by Bakumenko et al. [6]. This dataset, comprising anonymized journal entries, features system-specific account plans across multiple industries and timeframes. It has been originally preprocessed to exclude entries with more than four transactions to manage outliers. It includes a small subset of labeled anomalies with eight types of errors, created by financial auditors to reflect prevalent anomalies in financial records, indicating key areas of interest in real-world anomaly detection. The dataset focuses on attributes critical for anomaly detection, such as the source system, account category, and debit/credit indicators, streamlining the identification of irregularities within the data.

In this work, we applied rigorous ethical protocols to the General Ledger dataset, ensuring thorough anonymization to eliminate any identifiable details about companies or individuals. The dataset remains confidential and unshareable, safeguarding against unauthorized access. We avoided cloud storage to minimize data breach risks, maintaining the dataset's integrity. Data processing and analysis were conducted with strict adherence to legal and ethical guidelines. The introduced anomalies were carefully managed to uphold ethical data manipulation practices for research purposes.

## 4 METHODOLOGY

### 4.1 Data Preprocessing

The dataset comprises 32,100 transaction-level data points within journal entries, inclusive of 148 anomalies designed to reflect abnormal patterns without individual deviations. For anomaly detection in journal entries, transactions are aggregated into sets $J = \{T_1, T_2, \ldots, T_N\}$, where $J$ denotes a journal entry with $N$ transactions. An aggregated set $\mathcal{A}$ is formed by applying an aggregation function $A$ to each $J$, expressed as $\mathcal{A} = \{A(J) \mid J \in \mathcal{J}\}$.

In the work by Bakumenko Et al. [6], padding standardizes transaction lengths into uniform feature vector $F$, preparing ML model input. Transactions, defined by ERP attributes like account number and debit/credit sign, merged into the $ACCOUNT\_DC$ feature. The dimension of this encoded feature in a sparse matrix follows the formula:

$$F_{\text{count}} = \left( \max_{1 \leq k \leq K} \sum_{i=1}^{m_k} T_{k,i} \right) \cdot (|\mathcal{U}(V_{F1})| + |\mathcal{U}(V_{F2})|) \quad (1)$$

, where $F_{count}$ is the total feature count, calculated by the product of the maximum transaction amount across journal entries, denoted as $\max_{1 \leq k \leq K} \sum_{i=1}^{m_k} T_{k,i}$, and the combined count of unique elements in the $SOURCE$ and $ACCOUNT\_DC$ feature vectors ($V_{F1}$ and $V_{F2}$). Thus, on-hot encoding approach, where exist 577 unique

*ACCOUNT_DC* values and 4 unique values in the *SOURCE* feature, would result in 2336 encoded features. This feature space was PCA-reduced.

In contrast, to apply SBERT models for transactional data encoding in each JE, we first concatenated transactions categorical features with a group-by operation, based on JE identifiers. The procedure to combine the *SOURCE* and *ACCOUNT_DC* attributes of each transaction and the transactions themselves is:

$$C_i = \bigoplus_{j=1}^{n_i} \left(\text{"Source: "} + S_j + \text{" Account\_DC: "} + A_j\right) \quad (2)$$

, where $C_i$ is the concatenated text for the group $i$ and $n_i$ is the number of transactions in group $i$. $\bigoplus$ is the concatenation operation with a comma and space as transactions' delimiters. $S_j$ is the *SOURCE* attribute of the $j^{th}$ transaction in group $i$, and $A_j$ is the *ACCOUNT_DC* attribute of the $j^{th}$ transaction in group $i$.

The concatenated text ($Text_{conc}$) for each JE is processed as a single sentence structure. The SBERT model's encode method first tokenizes each string into a sequence of tokens. SBERT then generates contextual embeddings for each token using its BERT-based architecture, involving multiple transformer layers and self-attention mechanisms. A mean-pooling step, aggregates these token embeddings into a fixed-size sentence embedding. PCA-like dimensionality reduction isn't used to maintain the embedding's original dimensionality, ensuring precise evaluation [6]. We normalize embeddings to zero mean and unit variance for ML tasks to improve consistency and speed up convergence, crucial for distance-based or gradient descent algorithms, enhancing performance across models. SBERT embeddings create fixed-size dense vectors for each journal entry, capturing transaction details, aiding in anomaly detection and pattern recognition by summarizing complex data interactions.

## 4.2 Data Balancing and Model Performance Validation

In ML, skewed datasets with imbalanced class distribution hinder model training in classification tasks by favoring the majority class and affecting anomaly detection. Following the guidelines from [18], we used an 80/20 stratified split to ensure balanced training and testing sets with proportional anomaly representation, reducing bias. We adjusted for imbalance by weighting the minority class to improve sensitivity in model phases and ensured result consistency and fair comparison with a constant random state. In training and optimization, we avoided cross-validation, recognizing its shortcomings in imbalanced datasets and large feature sets, noted by Rao et al. [33]. Cross-validation raises overfitting risks, especially with many models and extensive hyperparameter tuning. Its efficacy drops as data dimensionality grows, causing higher model variance and unreliable assessments from intricate feature interactions. Imbalanced datasets exacerbate the challenge, leading to biased cross-validation folds and skewed performance evaluations.

We opted for a consistent 80/20 stratified split to maintain test set uniformity across models, crucial for accurately comparing algorithm performance, which cross-validation's variable data subsets could compromise. While this strategy mitigates some challenges, it potentially affects model generalizability. To counteract this, we employed careful metric selection and post-training cross-validation evaluation, although direct oversampling techniques for the minority class were impractical due to the dataset's complexity.

## 4.3 Model Selection

In this work, we evaluated three Sentence-BERT models for embedding generation: all-mpnet-base-v2, all-distilroberta-v1, and all-MiniLM-L6-v2, selected for their popularity and performance as indicated by their high download rates on the HuggingFace Model Hub [16]. Each model, trained on over 1 billion pairs, offers distinct advantages: all-mpnet-base-v2 excels in quality with a notable performance score of 63.30 [35], all-distilroberta-v1 provides a balance between efficiency and performance with a smaller size of 290 MB, and all-MiniLM-L6-v2 offers high speed with a compact size of 80 MB, suitable for real-time applications [35]. Refer to Table 1 for the detailed models' specifications. These models were chosen for their complementary strengths in quality, efficiency, and speed, facilitating a comprehensive evaluation in this research.

We also employed five ML classifiers: Random Forest (RF), Gradient Boosting Machines (GBM) using XGBoost (XGB), Support Vector Machines (SVM), Logistic Regression (LR), and Neural Networks (NN) implemented with Keras TensorFlow. RF is noted for its ability to reduce overfitting through ensemble decision trees, GBM for addressing data imbalance by optimizing weak learners, SVM for its effectiveness in high-dimensional spaces, LR as a fast and efficient baseline, and NN for modeling complex relationships, requiring careful architecture tuning [10, 12, 15, 17, 19, 23].

## 4.4 Experimental Design

Our work employed a financial dataset with both actual and artificially inserted anomalies, aiming to detect the latter while minimizing false positives among the former. This dataset, mirroring real-world conditions with significant class imbalance, is identical to that in Bakumenko et al.'s work [6], anonymized and refined to include only essential categorical features. We treated the 8 types of anomalies as a singular anomalous class, thereby framing it as a binary classification challenge, ensuring an even distribution of anomaly types in our train/test split.

Table 1: Specifications of Sentence-BERT Models [35]

| Specification | all-mpnet-base-v2 | all-distilroberta-v1 | all-MiniLM-L6-v2 |
|---|---|---|---|
| Base Model | microsoft/mpnet-base | distilroberta-base | nreimers/MiniLM-L6-H384-uncased |
| Dimensions | 768 | 768 | 384 |
| Size (MB) | 420 | 290 | 80 |
| Pooling | Mean Pooling | Mean Pooling | Mean Pooling |
| Average Performance | 63.30 | 59.84 | 58.80 |
| Speed | 2800 | 4000 | 14200 |
| Training Data | 1B+ sentence pairs | 1B+ sentence pairs | 1B+ sentence pairs |

To effectively identify anomalies in financial journal entries, which constitute a high-dimensional dataset, we innovated by encoding this non-semantic categorical data using SBERT LLMs, particularly employing 3 sentence-transformers models to standardize variable-length entries into a consistent feature space, addressing limitations in traditional encoding methods noted in the original work [6].

We explored five ML algorithms for anomaly detection, including Logistic Regression as a baseline, and optimized their hyperparameters using Hyperopt over 100 iterations. This fine-tuning, alongside the deployment of 3 Neural Network models with varied architectures, aimed to maximize model efficacy.

Model performance was assessed using the macro recall average metric, suitable for imbalanced datasets, focusing on the balanced detection of synthetic anomalies while controlling for false positives. This involved computing the average of Specificity and Sensitivity, alongside leveraging confusion matrices for a comprehensive performance overview [6].

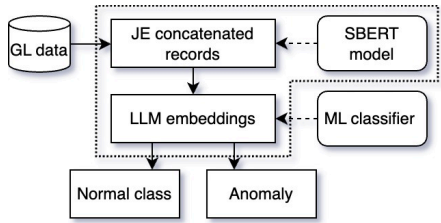

**Figure 2: High-level diagram of the anomaly detection algorithm with sentence-transformer and ML classifiers.**

Our experimental framework investigated the efficiency of LLMs in encoding financial data and assessed multiple ML models' anomaly detection capabilities. The anomaly detection system, as depicted in our high-level diagram in Figure 2, integrates LLM embeddings with an ML classifier to differentiate between normal and anomalous journal entries, starting from the General Ledger data, processed into aggregated records for ML classification based on LLM-derived embeddings.

## 5 EXPERIMENTATION RESULTS

### 5.1 Analysis of Encoded Feature Set

We conducted Principal Component Analysis (PCA) on feature sets derived from 3 SBERT model embeddings (Figure 3). PCA demonstrates the embeddings' dimensionality and information retention within the dataset, with significant variance preservation despite dimensionality reduction. The embeddings from the all-MiniLM-L6-v2 model (LLM1) required 63 components for 99% variance and 150 for 99.9%, whereas the all-distilroberta-v1 (LLM2) and all-mpnet-base-v2 (LLM3) models, despite larger vectors (770), needed fewer components (57 for LLM2 and 52 for LLM3) for the same variance level.

Further analysis revealed the less informative nature of the final 0.9% variance, suggesting it might contain noise or dataset-specific features. A comparative study (Figure 4) showed LLM embeddings' superior dimensionality reduction compared to one-hot encoding. The LLM embeddings maintained high variance with fewer dimensions, in contrast to the sparse, high-dimensional vectors from one-hot encoding. The one-hot encoded data initially had 2336 dimensions, reduced to 419 to achieve 99% variance, which is still higher than the LLM embeddings.

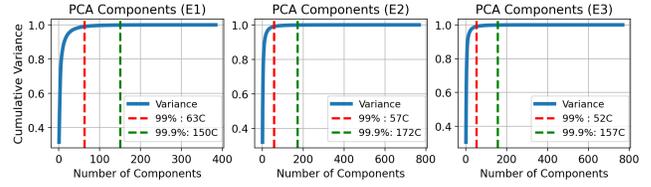

**Figure 3: PCA analysis of SBERT embeddings for the all-MiniLM-L6-v2 (E1), all-distilroberta-v1 (E2), and all-mpnet-base-v2 (E3) models. E1 achieves 99% variance with 63C and 99.9% variance with 150C. E2 achieves 99% variance with 57C and 99.9% variance with 172C. E3 achieves 99% variance with 52C and 99.9% variance with 157C.**

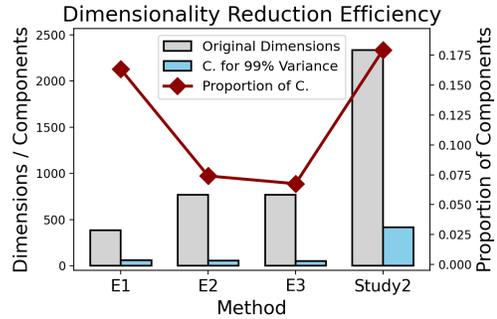

**Figure 4: Dimensionality reduction efficiency comparison between novel (E1: all-MiniLM-L6-v2, E2: all-distilroberta-v1, E3: all-mpnet-base-v2) and conventional method (Study2).**

It's important to recognize PCA's linear nature limiting its ability to capture non-linear complexities. While useful for understanding structural properties and potential for dimensionality reduction, PCA doesn't predict performance in downstream tasks. Our extended analysis includes empirical assessments of embeddings in these tasks.

In summary, LLM embeddings offer more efficient data representation than one-hot encoding, requiring fewer dimensions for similar variance levels, making LLM embeddings more preferable for complex tasks.

### 5.2 Downstream Model Training and Optimization

We utilized a variety of ML classifiers, as discussed in Section 4.3, including SVM, RF, XGBoost, LR, ANN, and DNNs. For non-ANN/DNN models, Bayesian optimization via the Hyperopt library and the Tree-structured Parzen Estimator (TPE) algorithm was applied over 100 iterations for hyperparameter tuning. Sample weights were calculated to address imbalanced datasets, and binary classification was achieved by transforming multi-class labels. Model training utilized Python with Scikit-learn and TensorFlow libraries. We designed three Neural Network architectures with varying complexities and implemented training over 50 epochs with early stopping for generalization, detailed in Table 2.

**Table 2: Configurations of Neural Network Models**

|  | ANN | DNN1 | DNN2 |
| --- | --- | --- | --- |
| Hidden layers | 1 | 3 | 3 + 2 Dropout |
| Neurons per layer | 64 | 256, 128, 64 | 256, 128, 64 |
| Dropout rate | - | - | 0.5 |
| Activation function | ReLU | ReLU | ReLU |
| Output activation | Sigmoid | Sigmoid | Sigmoid |

Reproducibility was ensured by fixing seeds in NumPy and TensorFlow, and recall average macro was monitored via custom callbacks. The ANN model featured a single hidden layer for rapid training, DNN1 had multiple hidden layers for complex pattern recognition, and DNN2 included dropout layers to prevent overfitting, maintaining a deep architecture like DNN1.

### 5.3 Evaluation Metrics and Comparative Analysis

Earlier in this work, we discussed benefits of having LR as a baseline model. For each of the three SBERT model embeddings, we trained two downstream LR models: a model with the default parameters and Hyperopt-optimized model. Non-optimised models showed high performance measured in recall average macro equal 0.9516, 0.9040, and 0.9520 for all-MiniLM-L6-v2, all-distilroberta-v1, and all-mpnet-base-v2 embeddings respectively. Figure 5 shows learning curves for optimized LR models, using a *learning_curve* function for cross-validation to check for generalization and overfitting. It trains the model on increasing data subsets and evaluates on training and validation sets, using 5-fold cross-validation. We calculated mean and standard deviation for training and validation scores across folds to gauge average performance and variability considering class imbalance. The training score line (red) indicates training subset performance, and the cross-validation score line (green) indicates unseen validation set performance, offering a robust estimate of model performance across data subsets and potential improvement with more data.

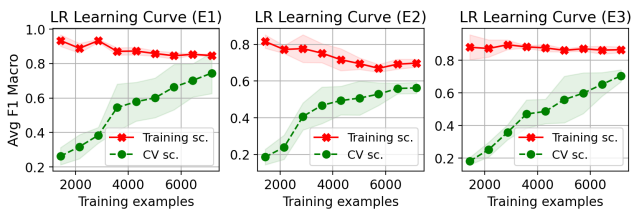

**Figure 5: Learning curve post-train evaluation for optimized LR models using all-MiniLM-L6-v2 (E1), all-distilroberta-v1 (E2), and all-mpnet-base-v2 (E3) model embeddings.**

In Figure 5, all three models demonstrate positive learning characteristics. The E3 model excels in learning and generalization, demonstrating strong data learning capacity. The E1 model, while performing adequately, shows signs of reaching its learning capacity limit. The E2 model is improving but requires better regularization strategies. The Hyperopt-optimized analysis shows E1's C value at 0.07677 indicates moderate regularization. E2 has a stronger regularization with a C value of 0.01702 and employs the 'newton-cg' solver. E3, with the smallest C value of 0.01358, exhibits the strongest regularization using the 'liblinear' solver. All models use uniform class weights to improve minority class prediction accuracy, with specific regularization strengths and solver selections tailored to their learning needs.

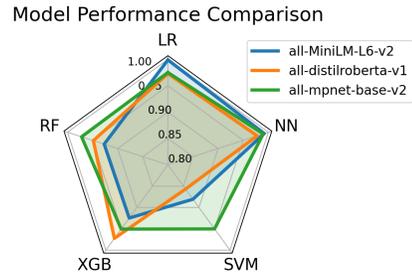

**Figure 6: Recall_AM scores for sentence-transformer and ML classifier models.**

Figure 6 evaluates the performance of LR, RF, XGB, SVM, and NN classifiers optimized and integrated with embeddings from three language models. Performance metrics are based on average recall macro. All embeddings show varying degrees of effectiveness, with all-mpnet-base-v2 excelling in stability and performance across various classifiers. all-MiniLM-L6-v2 also performs well, notably with LR and NN, while all-distilroberta-v1 is solid but does not outperform the all-MiniLM-L6-v2 with NN. These differences indicate that certain embeddings are more compatible with specific classifiers in downstream tasks, guiding practical model selection.

The confusion matrix values for the optimized models using different embeddings are provided in Table 3. The all-MiniLM-L6-v2 embeddings with LR, and potentially NN, if FPs are reduced, could provide the most balanced performance, while all-mpnet-base-v2 embeddings demonstrate higher TP rates across models. Figure 7 contrasts recall macro score differences for LLM embeddings compared to padded one-hot encoding across LR, RF, SVM, and NN models. Each bar shows the recall score difference for an LLM embedding, with boxplots summarizing the distribution and means (diamonds) for each model.

For LR, LLM embeddings improved recall scores by +0.056, +0.030, and +0.032, with a compact distribution indicating consistent enhancement across embeddings. In contrast, RF models showed a decline with LLM embeddings, marked by differences of -0.066, -0.044, and -0.021, and a moderate variability range. SVM models experienced reduced performance with LLM embeddings, with negative differences of -0.085, -0.108, and -0.015, showing significant variability towards lower performance. NN models benefited from LLM embeddings, with increases of +0.064, +0.050, and +0.062, and minimal variability, indicating a reliably positive effect. LLM embeddings improve LR and NN model performance over traditional padded one-hot encoding, but generally reduce effectiveness in RF and SVM models. While some models may consistently benefit from LLM embeddings, these highlights the model-specific variability in performance when applying LLM embeddings for data encoding.

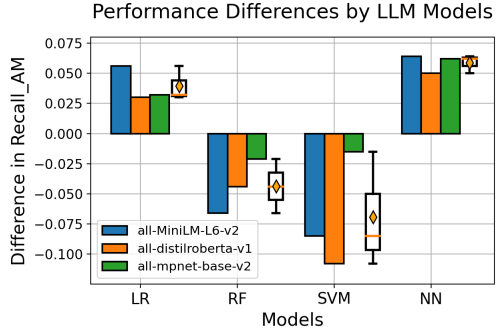

Figure 7: Performance differential analysis of ML models using LLM Embeddings versus traditional encoding.

The Bland-Altman plots in Figure 8 compare two score sets, evaluating ML model performances using embeddings (all-MiniLM-L6-v2 (E1), all-distilroberta-v1 (E2), and all-mpnet-base-v2 (E3)) against a traditional method. The red line shows the average difference in recall macro scores between all models. Blue lines, set at the mean difference ± 1.96 SD, define the limits of agreement, indicating the expected range for most score differences. Point dispersion around the blue lines shows some models' new embeddings align with expected performance ranges versus traditional methods. Performance varies across ML models and embeddings. For instance, NN models often exhibit improved results, shown by positive deviations above the red line, whereas SVM models display reduced efficacy, indicated by negative deviations. Overall, ML models employing LLM embeddings tend to match the anticipated performance spectrum of traditional methods, indicating on-average comparable outcomes.

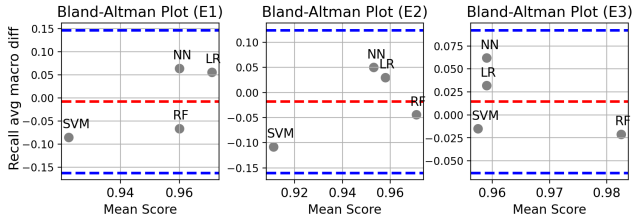

Figure 8: Model performance differentials using LLM Embeddings (all-MiniLM-L6-v2 (E1), all-distilroberta-v1 (E2), and all-mpnet-base-v2 (E3), compared to traditional encoding.

### 5.4 Hypotheses Revisited

The PCA analysis on sentence-transformer embeddings demonstrates improved compactness and information retention in financial data encoding over traditional approaches, confirming Hypothesis 1. This highlights the embeddings' superior ability to standardize feature variability and compress information effectively.

For Hypothesis 2, the integration of sentence-transformer embeddings with optimized LR and NN models, showed improved anomaly detection performance, affirming the LLMs embeddings potential to surpass traditional methods. Although some performance variances were observed, such as in SVM models, these were within anticipated bounds. The results underscore the efficacy of this innovative approach, emphasizing the importance of strategic model choice to maximize benefits.

## 6 DISCUSSION

This work's utilization of sentence-transformer LLMs for financial data encoding demonstrated a novel approach to enhancing anomaly detection.

### 6.1 Interpretation of Results

Using PCA on embeddings from three SBERT models (MiniLM-L6-v2, all-distilroberta-v1, and all-mpnet-base-v2) demonstrates a substantial improvement in dimensionality reduction and information retention for financial datasets compared to traditional encoding. For instance, considering downstream ML performance, the all-mpnet-base-v2 model needed just 52 PCA components to preserve 99% variance, compared to 419 for padded one-hot encoding. Embedded feature dimensionality for all 3 SBERT models was significantly lower in the same comparison. This advancement addresses the critical challenge of feature heterogeneity and sparsity in financial non-semantic non-temporal categorical feature sets, which is a notable improvement over traditional methods. Downstream ML models' performance confirms LLM embeddings' efficacy in anomaly detection. The employment of various ML classifiers, including Bayesian-optimized LR, RF, XGB, SVM, and NNs with multiple architectures and adjusted parameters, highlights the embeddings' versatility and potential to boost model performance. The superior evaluation metrics for LR and NN models using all 3 SBERT embeddings underscore these embeddings' potential in enhancing anomaly detection. The underperformance of SVM, even within the expected bounds, highlights the need for model-embedding compatibility assessment in future applications.

Table 3: ML model scores across various embeddings

| | all-MiniLM-L6-v2 | | | | | all-distilroberta-v1 | | | | | all-mpnet-base-v2 | | | | |
|---|---|---|---|---|---|---|---|---|---|---|---|---|---|---|---|
| Model | TN | FN | FP | TP | Recall_AM | TN | FN | FP | TP | Recall_AM | TN | FN | FP | TP | Recall_AM |
| Logistic Regression | 3809 | 0 | 8 | 21 | 0.9990 | 3792 | 1 | 25 | 20 | 0.9729 | 3808 | 1 | 9 | 20 | 0.9750 |
| Random Forest | 3805 | 3 | 12 | 18 | 0.9270 | 3788 | 2 | 29 | 19 | 0.9486 | 3785 | 1 | 32 | 20 | 0.9720 |
| XGBoost | 3788 | 3 | 29 | 18 | 0.9248 | 3789 | 1 | 28 | 20 | 0.9725 | 3804 | 2 | 13 | 19 | 0.9507 |
| Support Vector Machines | 3815 | 5 | 2 | 16 | 0.8807 | 3815 | 6 | 2 | 15 | 0.8569 | 3801 | 2 | 16 | 19 | 0.9503 |
| Neural Networks | 3756 | 0 | 61 | 21 | 0.9920 | 3648 | 0 | 169 | 21 | 0.9779 | 3740 | 0 | 77 | 21 | 0.9899 |

## 6.2 Implications for Financial Anomaly Detection

The integration of sentence-transformer LLMs in financial anomaly detection represents a leap from traditional methods, enhancing data representation and algorithm sensitivity to anomalies. This approach, in practical settings, promises to elevate fraud detection efficiency by improving accuracy and minimizing false alerts, thus streamlining financial operations. The novel method achieves an eightfold decrease in the number of components in certain scenarios while improving downstream model performance, efficiently standardazing feature variability. It illustrates its effectiveness and setting new standards in financial data encoding. As this methodology becomes more prevalent, it could establish new benchmarks in financial analysis, catalyzing advancements in ML applications within the industry. Further empirical studies and real-world applications could solidify its standing and quantify its impact.

## 6.3 Limitations and Bias

Our research utilized a real-world dataset from various ERPs, enriched with eight distinct types of intentionally introduced and labeled anomalies by financial auditors. The anomalies, reflecting auditors' interests in practical anomaly detection, have a synthetic nature that may limit generalizability. Additionally, the challenge in analyzing real-world financial data lies in unlabeled anomalies that may exist, potentially skewing ML model validation and increasing false-positive rates. Additionally, PCA analysis is a linear method limited in its capacity to represent non-linear relationships in feature sets. Also, LLMs respond to prompt engineering, meaning changes in input feature concatenation can alter embeddings, an aspect not covered in this work. Finally, our method focuses on categorical features and requires extension in scenarios requiring precise numerical analysis.

## 7 CONCLUSION AND FUTURE WORK

### 7.1 Contributions Summary

Our research advances the domain of financial anomaly detection through the integration of LLM embeddings with ML classifiers, a novel approach that notably mitigates feature heterogeneity and sparsity. Utilizing sentence-transformer models for financial data encoding, our methodology not only surpassed traditional encoding techniques in dimensionality reduction and information retention, but also showcased enhanced anomaly detection efficacy with selected ML classifiers. This aligns with established principles of feature representation [9], reflecting their practical application in the context of financial data. Underpinned by a comprehensive experimental setup and demonstrating practical applicability, our work contributes valuable insights for future research at the intersection of natural language processing and financial analytics.

### 7.2 Broader Impact and Implications

The innovative use of Large Language Models (LLMs) for non-semantic financial data tackles high-dimensionality and sparsity, establishes a precedent for the use of LLMs in domains beyond their traditional applications. This mirrors findings where LLMs successfully encode visual tokens [30]. By outperforming conventional methods, LLM embeddings show their potential beyond linguistic tasks, particularly for data types lacking inherent semantics. This methodological advancement could aid various industries with similar challenges, notably healthcare and retail, where complex datasets might gain from the enhanced data representation capabilities of LLMs. In healthcare, LLM embeddings could enhance patient data analysis by detecting patterns in datasets that are mainly numerical, lack textual clarity, or consist of structured data like MRIs, CT scans, ICD codes, and lab values, which require domain knowledge for interpretation. In retail, LLMs could offer detailed insights from high-dimensional transactional data, revealing complex product-consumer interactions. This can enhance ML models' ability to predict behaviors, segment markets, and suggest products, advancing market analytics.

The utilization of LLMs on non-semantic data expands their use and prompts a rethinking of data analysis methods, fostering multidisciplinary research into their potential for complex datasets.

### 7.3 Future Research Directions

Future research should extend the LLM embedding approach to broader financial datasets, assessing its scalability, impact on anomaly detection accuracy, and computational efficiency in response to evolving financial fraud patterns. Extending this methodology to various non-semantic data types across multiple domains with high-dimensional and sparse datasets, and integrating with other advanced ML and deep learning models, will test the adaptability and effectiveness of LLM embeddings. Unsupervised strategies should be explored to address zero-day anomalies, refining our method to better detect novel patterns. Future research should investigate how diverse data preprocessing strategies, including aggregation methods and prompt engineering, enhance LLM encoding efficiency. A focused exploration into non-linear dimensionality reduction techniques could complement PCA, aiming to more effectively capture complex relationships within LLM embeddings. Investigating the effects of synthetic versus real-world anomalies on model performance will offer insights into the findings' practical applicability. Finally, exploring model-embedding compatibility by testing various cutting-edge LLM architectures could yield more tailored anomaly detection solutions.